\documentclass[letterpaper, 10 pt, conference]{ieeeconf}  

\IEEEoverridecommandlockouts                              

\usepackage{graphics} 
\usepackage{epsfig} 
\usepackage{mathptmx} 
\usepackage{times} 
\usepackage{amsmath} 
\usepackage{amssymb}  

\usepackage{amsfonts}
\usepackage{bm}
\usepackage{makecell}
\usepackage{dsfont}
\usepackage{booktabs} 
\usepackage[linesnumbered,ruled,norelsize]{algorithm2e}
\usepackage{hyperref}
\newcount\Comments 
\usepackage{array,multirow,graphicx}
\usepackage{color}
\usepackage{wrapfig}
\definecolor{darkgreen}{rgb}{0,0.5,0}
\definecolor{purple}{rgb}{1,0,1}
\newcommand{\kibitz}[2]{\ifnum\Comments=1\textcolor{#1}{#2}\fi}

\usepackage{stfloats}
\usepackage{tabularx}
\let\labelindent\relax
\usepackage{enumitem}
\usepackage{subcaption}
\usepackage{graphicx}
\usepackage{siunitx}
\usepackage{float}

\title{\LARGE \bf
Learning Your Way Without Map or Compass: Panoramic Target Driven Visual Navigation
}

\author{David Watkins-Valls$^{*,1}$,
        Jingxi Xu$^{*,1}$,
        Nicholas Waytowich$^2$ and
        Peter Allen$^1$
\thanks{$^{*}$Authors have contributed equally and names are in alphabetical order.} 
\thanks{$^1$Department of Computer Science, Columbia University, New York, NY, USA. {\tt\small\{davidwatkins, allen\}@cs.columbia.edu}, {\tt\small jingxi.xu@columbia.edu}}
\thanks{$^2$U.S. Army Research Laboratory, Baltimore, MD, USA. {\tt\small nicholas.r.waytowich.civ@mail.mil}}
\thanks{This work is supported by NSF Grant CMMI 1734557. This research was sponsored by the Army Research Laboratory and was accomplished under Cooperative Agreement Number W911NF-18-2-0244. The views and conclusions contained in this document are those of the authors and should not be interpreted as representing the official policies, either expressed or implied, of the Army Research Laboratory or the U.S. Government. The U.S. Government is authorized to reproduce and distribute reprints for Government purposes notwithstanding any copyright notation herein.}
\thanks{The title is an homage to Harold Gatty's book \textit{Finding Your Way Without Map or Compass}.}
}

\begin{document}

\maketitle
\thispagestyle{empty}
\pagestyle{empty}


\begin{abstract}
We present a robot navigation system that uses an imitation learning framework to successfully navigate in complex environments. Our framework takes a pre-built 3D scan of a real environment and trains an agent from pre-generated expert trajectories to navigate to any position given a panoramic view of the goal and the current visual input without relying on map, compass, odometry, or relative position of the target at runtime. Our end-to-end trained agent uses RGB and depth (RGBD) information and can handle large environments (up to $1031m^2$) across multiple rooms (up to $40$) and generalizes to unseen targets. We show that when compared to several baselines our method (1) requires fewer training examples and less training time, (2) reaches the goal location with higher accuracy, and (3) produces better solutions with shorter paths for long-range navigation tasks.

\end{abstract}

\section{Introduction}
The ability to navigate efficiently and accurately within an environment is fundamental to intelligent behavior and has been a focus of research in robotics for many years. Traditionally, robotic navigation is solved using model-based methods with an explicit focus on position inference and mapping, such as Simultaneous Localization and Mapping (SLAM)~\cite{dissanayake2001solution}. These models use path planning algorithms, such as Probabilistic Roadmaps (PRM)~\cite{kavraki1994probabilistic} and Rapidly Exploring Random Trees (RRT)~\cite{lavalle2000rapidly, kuffner2000rrt}, to plan a collision-free path. These methods ignore the rich information from visual input and are highly sensitive to robot odometry and noise in sensor data. For example, a robot navigating through a room may lose track of its position due to the navigation software not properly modeling friction. 

Model-free reinforcement learning (RL) agents have performed well on many robotic tasks~\cite{kober2013reinforcement, mulling2011biomimetic, mnih2015human, andrychowicz2017hindsight}, leading researchers to rely on RL for robotic navigation tasks~\cite{THOR,francis2019long, COMPLEX_ENVS, faust2018prm}.
Recent work in robotic visual navigation uses reinforcement learning which trains an agent to navigate to a goal using only the current and goal RGB images~\cite{THOR}. While reinforcement learning has the convenience of being weakly supervised, it generally suffers from sparse rewards in navigation, requires a huge number of training episodes to converge, and struggles to generalize to unseen targets. The problem is further exacerbated when the navigation environment becomes large and complex (across multiple rooms and scenes with various obstacles), leading to difficult long-range path solutions.  


\begin{figure}[t]
    \centering {
        \includegraphics[width=0.9\linewidth]{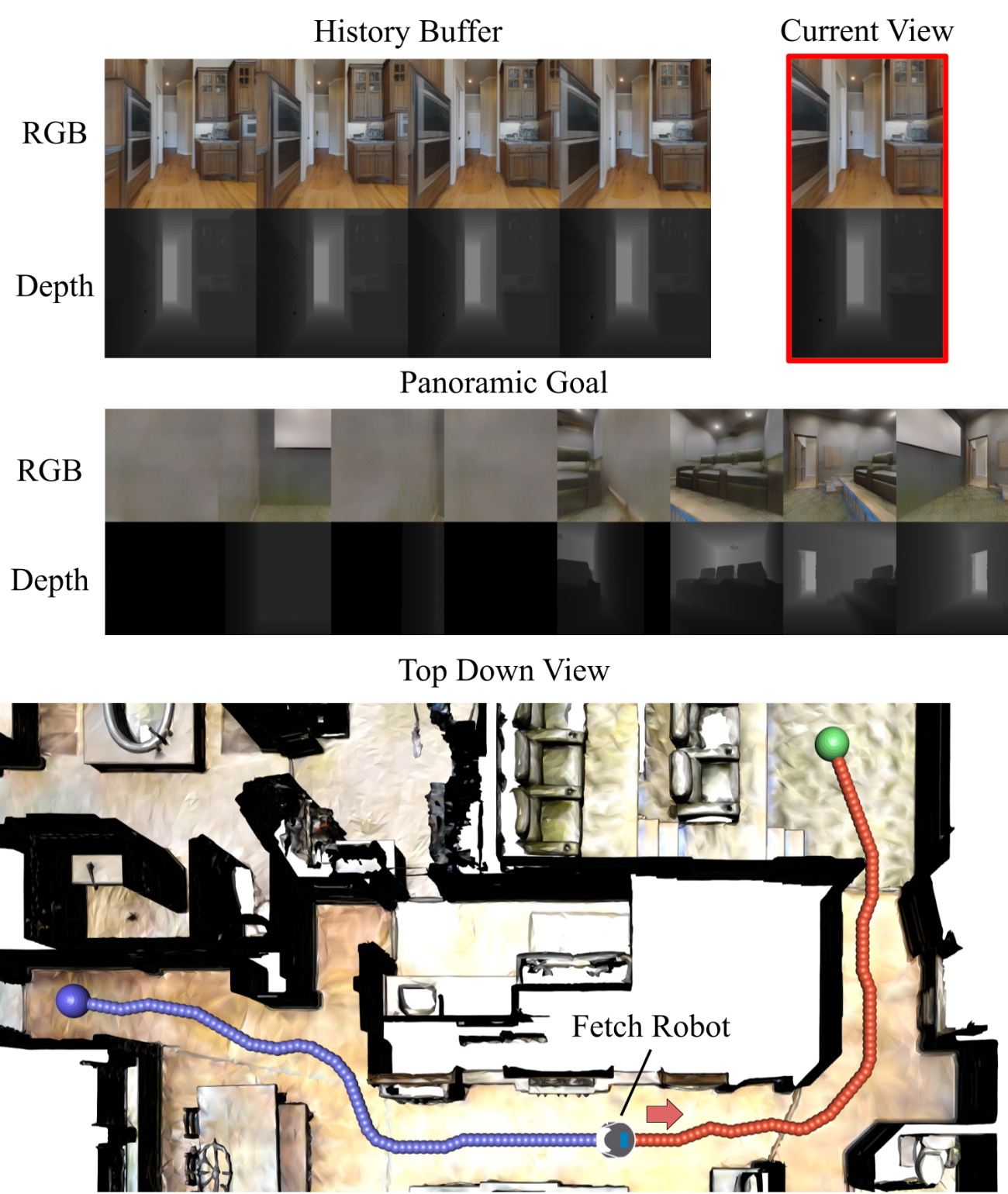}
    }
    \caption{A successful trajectory executed in \texttt{house17} from the Matterport3D dataset. The history buffer and current view are the state of the pipeline. The panoramic goal is 8 RGBD images each taken at a \ang{45} turn. The top-down view is the agent moving through the trajectory with the blue sphere as the start position and the green sphere as the goal position. Smaller blue spheres are positions that the agent has been to and the orange spheres are the remaining positions. The images are taken at the current position of the robotic agent. }\label{fig:title_picture}
    \vspace{-4mm}
\end{figure}

New advancements in annotated 3D maps of real-world data, such as Stanford2D3DS~\cite{stanford2d3ds} and Matterport3D~\cite{Matterport3D}, enable the collection of large amounts of trajectory data. Simulators capable of collecting this data have arisen in the past few years in the form of MINOS~\cite{minos}, Gibson~\cite{GIBSON}, Habitat~\cite{habitat}, and THOR~\cite{THOR}. These systems enable simultaneous use of real and simulated environments for training, without the need for visual domain adaptation. Gibson~\cite{GIBSON} features real-world and photo-realistic data generated from fully scanned 3D homes and buildings that allow for easier collection of demonstration data for supervised learning.

Our work focuses on exploring supervised methods (in particular, imitation learning) to bring better performance to robotic visual navigation, while taking advantage of the current progress in robotic simulators and datasets to efficiently collect training data and handle the sim-to-real gap and domain adaptation. Our contributions are as follows: (1) we provide a navigation pipeline where the agent learns to navigate to unseen targets using the current RGBD view and a novel 8-image panoramic goal without using compass, map, or relative position of goals at runtime; (2) we provide a framework to efficiently generate optimal expert trajectories in the Gibson simulator using a 3D scan of the environment of interest; and (3) we provide a methodology for discretizing a continuous trajectory into a series of \{\textit{forward}, \textit{right}, \textit{left}\} commands. Our method outperforms alternative methods in both the quality of the solution paths and the success rate, with significantly fewer training examples and less training time. A longer video, dataset, and source code can be found at \url{http://crlab.cs.columbia.edu/learning_your_way/}.

\section{Related Work}
\paragraph{Reinforcement learning methods for navigation} 
Previous work in visual navigation~\cite{THOR} provides a target-driven reinforcement learning framework for robotic visual navigation. Our method shares the same objective of navigating to the goal position using the goal image, the current image, and a sequence of history images. \cite{splmetric} and \cite{habitat} train an RL agent to navigate in realistic cluttered environments using a PointGoal (a specific location for the goal target). They assume an idealized GPS which provides the relative goal position and use this information to train their agents. Both \cite{habitat} and \cite{THOR} claim that their learned policy generalizes across targets and environments. \cite{THOR} only evaluates their method on new targets that are several steps away from the targets that the agent is trained on and the scene-specific layer has to be retrained for the policy to work in a new environment. \cite{habitat} relies on an idealized GPS and the specific location of the goal and it generalizes to new environments by learning a bug algorithm like behavior to follow the boundaries. Imagine a scenario in which a person is placed into a building they have not seen before with nothing but an image of the place they need to get to. It would be unfair for us to expect this person to navigate to the target location in any efficient manner. Therefore, a robotic agent would be unable to handle generalizing to new untrained environments using vision alone and we focus on the ability to generalize to any unseen target in the environment that the robotic agent has seen before.

Previous work~\cite{COMPLEX_ENVS} uses a synthetic 3D maze environment and the agent is trained on a single goal. Another work~\cite{mirowski2018learning} trains an agent to navigate using real-world Google Map views with the goal coordinate provided. \cite{faust2018prm, francis2019long, chiang2019learning} present hierarchical robot navigation methods using reinforcement learning to learn local and short-range obstacle avoidance tasks and using sampling-based path planning algorithms as global planners. These methods use 1D lidar sensor data and a dynamic goal position as input. \cite{gupta2017cognitive, khan2017memory} use value iteration networks~\cite{tamar2016value} to learn navigation strategies in simplistic synthetic simulated environments. \cite{mousavian2019visual} evaluates different representations for target-driven visual navigation using a semantic target and an off-the-shelf segmenter. \cite{bruce2017one} presents a method with an interactive world model to navigate to a fixed goal in a known environment. \cite{kahn2018self} reconstructs a navigation graph of an agent moving through an environment; however, it does not rely on RGBD alone. \cite{zhi2019learning} uses a novel methodology for mapping an environment using semantic information but does not solve navigation to an RGBD goal in their method. None of these works solve the problem of indoor visual navigation because they either make compromises in sensory input (1D lidar, goal position) or have a different environmental design (synthetic maps, outdoor data).

\paragraph{Supervised learning methods for navigation}
Because most work using learning methods for robotic navigation relies on deep RL, supervised methods are less explored. \cite{richter2017safe, lind2018deep} uses Convolutional Neural Networks (CNNs) for robotic navigation. However, their methods are different from ours in that they use odometry and a goal location as input into their networks and their models are trained only for simple tasks such as collision detection and localization.

Other work uses a 2D cartesian space to learn a function to imitate for finding an optimal path in an environment \cite{bency2019neural}\cite{ichter2017learning}\cite{ho2016generative}, however these works do not consider the problem of navigation in a 3D context and limit optimization within a 2D space or treat the robot as a point agent. 

\paragraph{Datasets and simulators for navigation environments}
As the broader area of active and embodied perception has received increased interest, new datasets (Stanford2D3DS~\cite{stanford2d3ds}, Matterport3D~\cite{Matterport3D}, SUNCG~\cite{suncg}, Gibson~\cite{GIBSON}) and simulators (MINOS~\cite{minos}, Gibson~\cite{GIBSON}, Habitat~\cite{habitat}, AI2THOR~\cite{kolve2017ai2}) have been created for robotic navigation. These new environments enable researchers to train an agent in simulation using real-world data and obtain training data much faster than would be possible in the real-world alone. They allow agents to be trained at scale using ground truth positioning and fast rendering. The Gibson simulator uses PyBullet~\cite{PYBULLET} to simulate collisions with the environment as well as dynamic environment tasks.

\section{Problem Setup}
The goal of this work is to enable the robot to autonomously navigate to a target position, described by a set of panoramic images taken at the goal, without providing any odometry, or the relative indoor location of the target but only RGBD input from the robot's point of view. The agent is aligned with one of the images in the panoramic goal once it has arrived at the final location using a local plan. We find the most likely similar image in the panoramic goal and compare with the current view of the agent, and then turn that many degrees until we face the correct direction in the local planning step. The problem is referred to as \textit{target-driven visual navigation} in the literature~\cite{THOR}, where the task objective (i.e., navigation destination) is specified as inputs to our model. Traditional learning-based visual navigation methods have largely focused on learning goal-specific models that tackle individual tasks in isolation, where the goal information is hard-coded in the neural network representations, leading to poor generalization to unseen\,/\,unexplored targets. Target-driven approaches learn to navigate to new targets without re-training, using a single navigation pipeline. 

Our navigation pipeline, denoted as $\Pi$, takes as input the observation of the current state $s_i$ at time step $i$, the target information $g$, and outputs an action $a_i \in \{forward, right, left, done\}$.

\begin{equation}
    a_i = \Pi \left( s_i, g \right)
\end{equation}

The \textit{left}\,/\,\textit{right} action indicates turning the agent in place left\,/\,right 10 degrees and the \textit{forward} action moves the agent 0.1$m$ ahead. The unknown transition model $\Gamma$ of the environment updates the state, denoted by $s_{i+1} = \Gamma(s_i, a_i)$, when an action $a_i$ is executed. The objective is that given any goal $g$ in the map, a maximum number of steps $T$, and a success threshold $\zeta$, our navigation pipeline $\Pi$ can generate a sequence of actions $\{a_i\}, i\in[t]$, which satisfies (1) $t < T$; (2) $a_t = done$; (3) the final location of the robot is within $\zeta$ meter of the target location; and (4) the length of the path should be as short as possible. Our navigation pipeline is fully automated as it learns to stop at the goal and does not require human intervention through the whole process. We choose 0.1$m$ and 10 degrees because they allow for the correct discretization of the original planned path within an error of 0.01$m$ of the goal location. We could choose smaller values, but instead we strike a good balance between time to plan and accuracy of the approximated path. 

The state $s_i$ is the current RGBD visual observation and a history buffer of 4 concatenated past RGBD images, both of which are from the agent's viewpoint. The goal information $g$ is a set of 8 panoramic RGBD images. An example of the state and the goal information is shown in Figure~\ref{fig:title_picture}.

\section{Navigation Pipeline}
\subsection{Overview}
Our navigation pipeline $\Pi$ consists of three separately trained models using neural networks, the autoencoder model $A$, the policy model $E$, and the goal checking model $G$.  

\begin{figure}[h!]
    \vspace{2mm}
    \centering {
        \includegraphics[width=\linewidth]{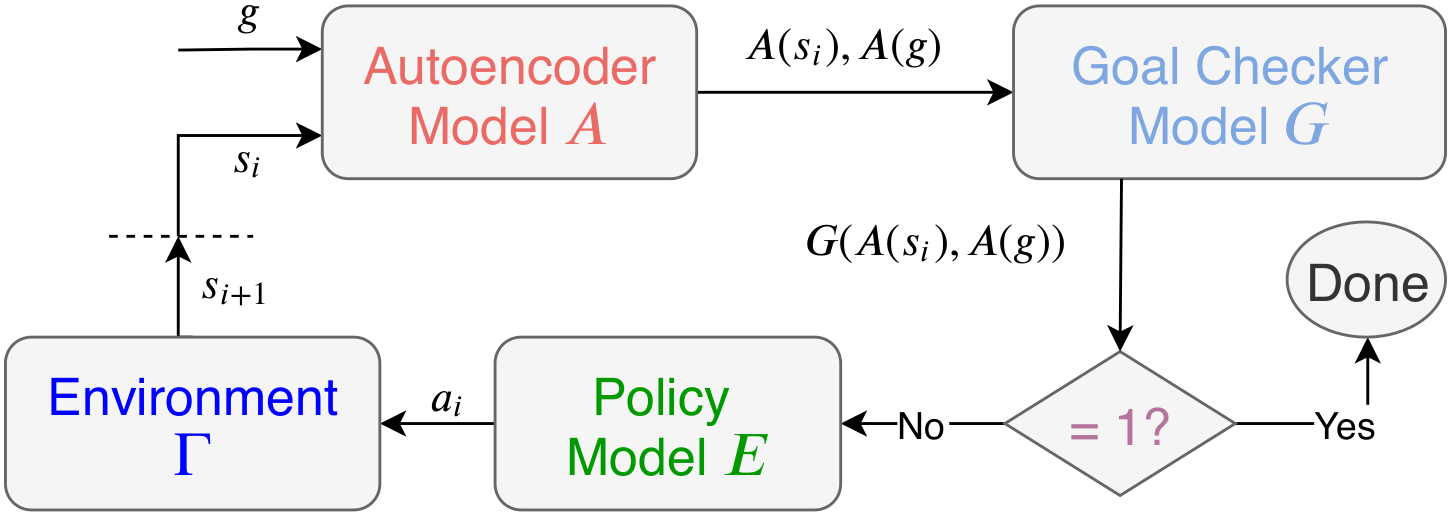}
    }
    \caption{Overview of our navigation pipeline architecture. The flow starts when a new navigation task is received, and continues until a $done$ action is generated.}
    \label{fig:overview}
\end{figure}

The autoencoder model generates latent representations (i.e., embeddings) for both the state $s_i$ and the goal $g$, denoted $A(s_i)$ and $A(g)$. The policy model takes two inputs, the embeddings of the current state and the embeddings of the target, and produces a probability distribution over three actions, $\displaystyle a_i \in \{\textrm{\textit{forward}, \textit{right}, \textit{left}}\} \sim E\left(A(s_i), A(g) \right))$. It then picks the action with highest probability from this distribution. The policy model is responsible for leading the agent towards the goal with as little exploration as possible. The goal checking model is a binary function which takes the same input as the policy model, and decides if the agent has reached the target or not, denoted by $G(A(s_i), A(g)) \in \{1, 0\}$, where 1 corresponds to \textit{done} and 0 corresponds to \textit{not done}. An overview of the navigation pipeline is shown in Figure~\ref{fig:overview}.

\subsection{Autoencoder Model}
Because the input into our neural network models is RGBD images, the training is more efficient if we use embeddings of the raw input. Instead of extracting features from an intermediate layer of a pre-trained classifier such as ResNet-50~\cite{THOR, he2016deep}, we train an autoencoder from images captured from the same environment.

\begin{figure}[h!]
    \centering {
        \includegraphics[width=0.85\linewidth]{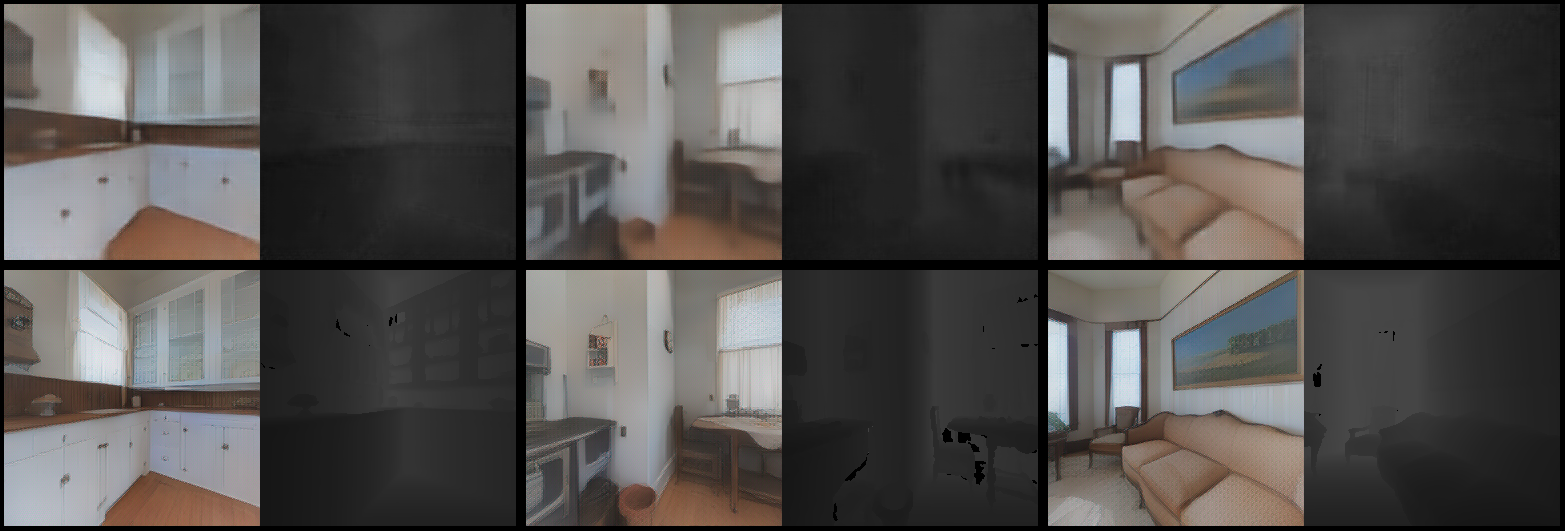}
    }
    \caption{An example of reconstructed images from the autoencoder model trained in the \texttt{house2} environment. The top row is 3 predicted output images (RGB image appended by depth image); the bottom row is the original images.}\label{fig:autoencoder_imgs}
\end{figure}

Similar to RedNet~\cite{jiang2018rednet}, our autoencoder network is based on a 6-layer CNN with batch normalization on every layer. The reconstruction half of the network is made up of an additional 6 transposed convolutional layers with batch normalization applied before each transposed convolution. We use rectified linear unit (ReLU) as the activation function and Adam optimizer~\cite{kingma2014adam} to minimize the mean squared error between the reconstructed and the original images. The autoencoder is able to compress a $256\times256\times4$ RGBD image into the $4096$-d latent space ($\times 64$ space savings). It is then used to encode each image of the state and each image of the panoramic goal. A detailed topology of the encoder section is pictured in Figure~\ref{net:autoencoder}. An example of the autoencoder performance is shown in Figure~\ref{fig:autoencoder_imgs}.

\subsection{Policy Model}
The policy model takes as input the embeddings of stacked observations and the panoramic goal images to generate the next action $a_i \in \{\textrm{\textit{forward}, \textit{right}, \textit{left}}\}$. We use imitation learning to teach the model how to navigate in a given environment.

Our policy model is a fully-connected multilayer perceptron (MLP) as shown in Figure~\ref{net:policy}. We also evaluated the performance of a variety of other deep learning architectures including convolution along the temporal dimension and long short-term memory (LSTM)~\cite{hochreiter1997long},  with different numbers of past images in the state and different numbers of panoramic goal images, but the MLP architecture with 4 past images and 1 current image outperforms the others. Its larger number of parameters increases its ability to model complex functions.  

The embeddings of the state and the panoramic goal are first concatenated to form a $13 \times 4096$ matrix and then progress through 3 fully-connected layers followed by batch normalization and ReLU activation after each layer, to generate a 16-d vector. The 16-d vector passes through the last fully-connected layer to generate 3 logits. Softmax activation then outputs a distribution over 3 actions \{\textit{forward}, \textit{right}, \textit{left}\}. We use Adam optimizer on the cross-entropy loss for back propagation. At testing, we pick the action with highest probability deterministically.

\begin{table}
    \centering {
    \vspace{4mm}

        \resizebox{\columnwidth}{!}{%
            \begin{tabular}{ccccc}
                \toprule
                \textbf{Model} & \textit{Success rate} & \textit{SPL} & \textit{Number of actions} & \textit{Timeout rate}\\
                \hline
                \textit{1H+1G} & 0.8847 & 5.212 & 524.003 & 0.1153 \\
                \textit{5H+1G} & 0.8925 & 3.779 & 478.401 & 0.1075 \\
                \textit{\textbf{5H+8G}} & \textbf{0.9725} & \textbf{2.322} & \textbf{274.846} & \textbf{0.0275} \\
                \textit{5H+8G+2S} & 0.8975 & 2.486 & 288.162 & 0.1025 \\
                \textit{5H+8G+conv} & \textbf{0.9725} & 2.865 & 340.884 & \textbf{0.0275} \\
                \textit{15H+8G+conv} & 0.945 & 4.412 & 476.219 & 0.055 \\
                \textit{15H+8G+LSTM} & 0.8125 & 6.564 & 685.395 & 0.1875 \\
                \textit{25H+8G+conv} & 0.9025 & 4.652 & 556.202 & 0.0975 \\
                \hline
            \end{tabular}
        }
    }
    \caption{Ablation study statistics related to the policy model during training. We train seven model architectures each of varying image input sizes and training strategies. We find that the model \textit{5H+8G}, with 5 input images and an 8-image panoramic goal, performs best at navigating to the goal location. Each of these models are evaluated using the ground truth goal checker with known indoor position. SPL is explained in Section \ref{sec:metrics}.}
    \label{table:ablation_study}
\end{table}

As an ablation study, we evaluate the performance of several deep learning architectures on a subset of the \texttt{area1} environment in the Stanford2D3DS dataset~\cite{stanford2d3ds} and pick the model which performs best for all experiments described in Section~\ref{sec:experiments}. The models are:
\subsubsection{No history and 1-image goal} The model is provided only the current view and 1 image of the goal. (See 1H+1G in Table~\ref{table:ablation_study})
\subsubsection{History buffer and 1-image goal} We use the 4 previous image embeddings that the agent saw in the trajectory in addition to the current image and 1 image for the goal. This is the architecture~\cite{THOR} used except we used a custom image embedding procedure instead of ResNet50~\cite{he2016deep}. (See 5H+1G in Table \ref{table:ablation_study}) 
\subsubsection{History buffer and panoramic goal} The final system architecture we use. It uses the 4 previous image embeddings, the current image embedding, and an 8-image panoramic goal embedding as input. It outperforms all other methods. We also tested with training every third image. (See 5H+8G and 5H+8G+2S in Table~\ref{table:ablation_study})
\subsubsection{History buffer and panoramic goal with convolution} Similar to 3D convolution~\cite{tran2015learning} on image sequences, we convolve over the input embeddings to allow for more input images to fit in a model during training. We test with 5, 15, and 25 input images. (See 5H+8G+conv, 15H+8G+conv, and 25H+8G+conv in Table~\ref{table:ablation_study})
\subsubsection{LSTM} We test an LSTM architecture as well which takes in the previous 14 image embeddings, the current image embedding, and 8 goal image embeddings. (See 15H+8G+LSTM in Table~\ref{table:ablation_study})

Overall, we find that the \textit{5H+8G} network outperforms all other model architectures in our test suite. We believe the \textit{convolution} models are unable to sufficiently learn a policy despite receiving more data as input due to the loss of data by replacing a fully connected layer with a convolutional layer. The \textit{LSTM} model is not able to sufficiently learn a policy that would have generalized to the full layout of the environment due to its lower accuracy and much longer paths. Providing the agent with 1 goal image is not enough to sufficiently learn a policy either and it is outperformed by the models which received 8 images for the goal. All results for this ablation study are shown in Table \ref{table:ablation_study}.

\begin{figure}
\centering
\begin{subfigure}[b]{0.45\textwidth}
   \vspace{2mm}
   \includegraphics[width=1\linewidth]{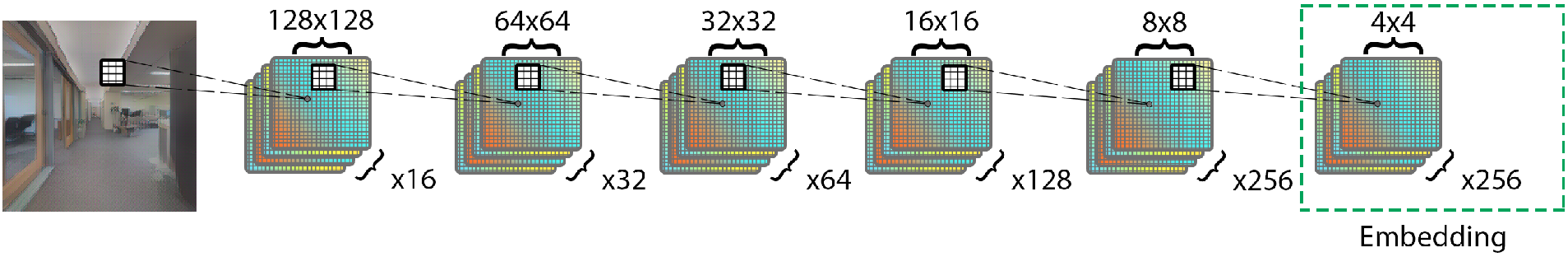}
   \caption{}
   \label{net:autoencoder} 
\end{subfigure}

\begin{subfigure}[b]{0.45\textwidth}
   \includegraphics[width=1\linewidth]{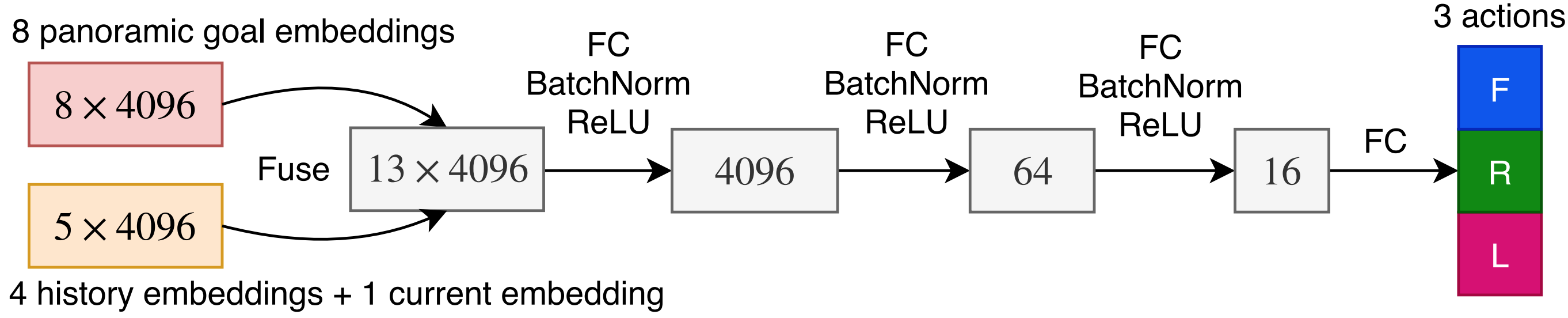}
   \caption{}
   \label{net:policy}
\end{subfigure}
\par\bigskip

\begin{subfigure}[b]{0.45\textwidth}
   \includegraphics[width=1\linewidth]{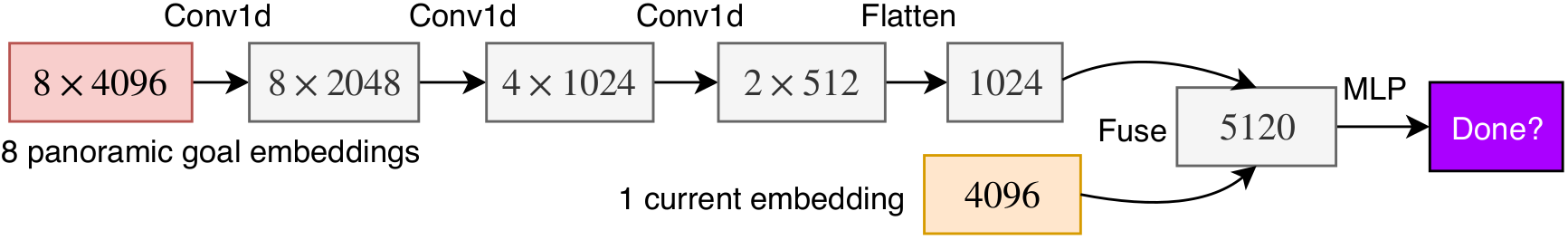}
   \caption{}
   \label{net:goal} 
\end{subfigure}

\caption{(a) Encoder architecture of our autoencoder. Progression through each layer consists of a convolution with a stride of 2 followed by batch normalization and ReLU activation. (b) Policy model architecture. \texttt{Fuse}, \texttt{FC}, and \texttt{Flatten} represent the concatenation operation, a fully-connected layer, and the flatten operation respectively. (c) Goal checking model architecture. \texttt{Conv1d} is the 1D convolution operation.}
\end{figure}

\subsection{Goal Checking Model}
The goal checking model takes in the embeddings of the current observation concatenated with the panoramic goal images and predicts whether the agent is at the target position, as shown in Figure~\ref{net:goal}. 

This model is created in response to an optimization on our original architecture which had the policy model output a \textit{done} action when the robotic agent arrives at a goal position. We find that the training data is too sparse for the agent to effectively learn identifying a goal location because we only have one positive example of \textit{done} at the end of each trajectory. All the other steps are negative examples for \textit{not done}. There is a significant imbalance in the number of positive and negative examples. In addition, at runtime our robot is likely to arrive at the target position from a different viewpoint than those in the panoramic goal images, but during training the policy model receives a view that is one of the panoramic goal images. Thus, we implement an additional binary classifier to identify whether the agent has arrived at the goal location. When this model predicts a \textit{done} action the navigation pipeline terminates.

The goal checking model is a dual-branch network with 1D convolution over the panoramic goal branch. The 1024-d vector from the goal-branch is then concatenated with the current embedding branch to form a 5120-d fused vector, which then passes through an MLP with a hidden layer of 512 units to output the probability of the goal being reached. Weights are then updated using an Adam optimizer on the cross-entropy loss. While using the learned goal checker at runtime, in order to reduce noise, the agent does a \ang{360} rotation when its belief of reaching the goal is over 0.99. It calls the learned goal checker after each \ang{10} turn. The agent outputs \textit{done} only if the average probability is over 0.9.

\section{Experiments}\label{sec:experiments}
\begin{figure*}[t]
\vspace{2mm}
    \centering {
        \includegraphics[width=0.95\textwidth]{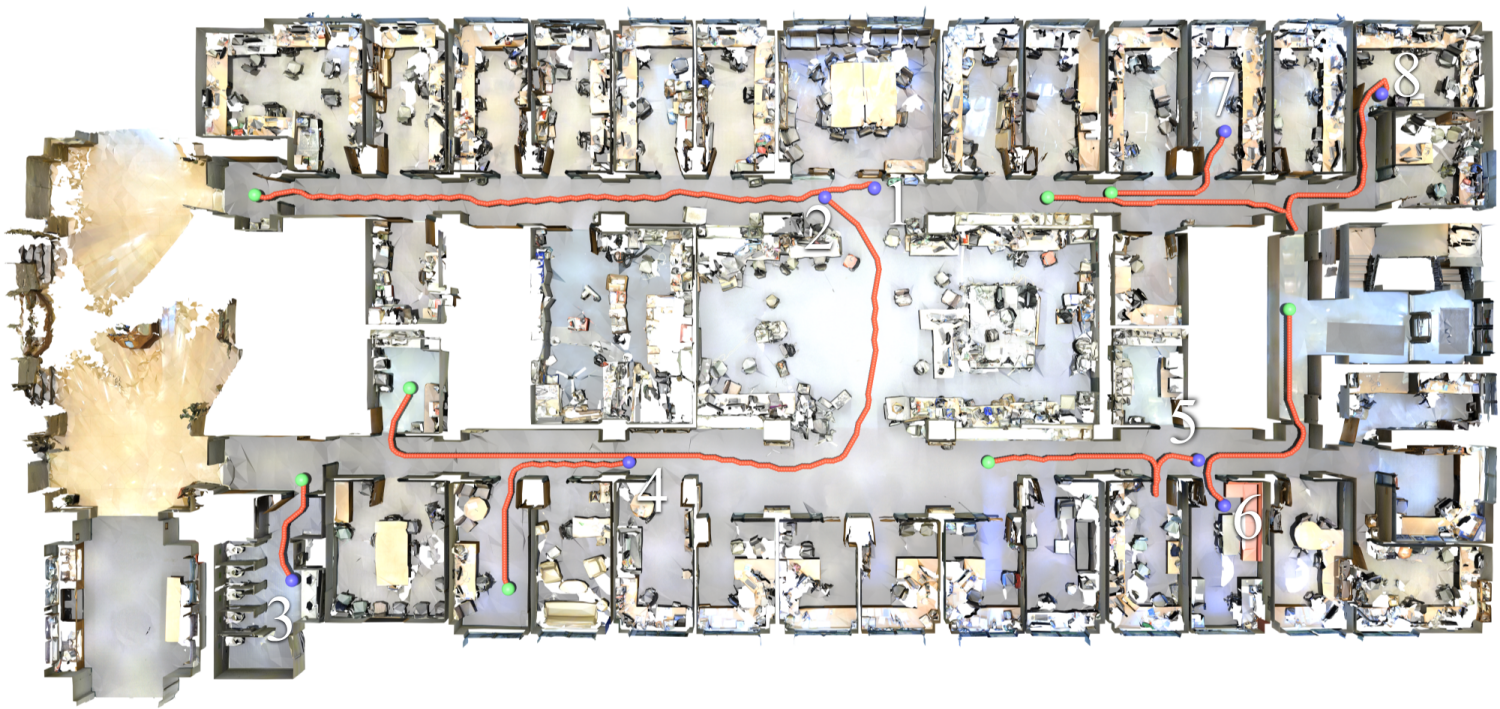}
    }
    \caption{Eight randomly selected non-overlapping successful trajectories in \texttt{area1}. Blue dots are start positions and green dots are goal positions. Trajectory $5$ and $8$ show recovery behavior which ultimately leads to a successful trajectory.} \label{fig:area1_demo}
    \vspace{-2mm}
\end{figure*}

We evaluate our navigation pipeline in 2 environments selected from the Stanford2D3DS dataset and 3 environments selected from the Matterport3D dataset. Metadata (including number of rooms and area) are shown in Table~\ref{table:our_performance}. We use the Gibson simulator with a Fetch~\cite{fetch} robot and focus on how the navigation pipeline generalizes to unseen targets under the same trained environment. All experiments are conducted using an NVIDIA 1080Ti GPU. Examples of planned paths, recovery behavior, and experimental environments can be found in the attached video.

\subsection{Network Training Setup}
For each submodule (autoencoder, policy, and goal checker) in our navigation pipeline, we need to generate the corresponding dataset for training and testing using the colored meshes loaded in the Gibson simulator. We pick the model with least loss (autoencoder) or highest accuracy (policy, goal checker) on the test set. Unless otherwise specified, we use a learning rate of 0.001.

\subsubsection{2D Environment Map}
We create a high resolution 2D map of the environment from its colored mesh, which enables trajectory planning using Dijkstra's algorithm. Each of the maps uses a resolution of $1cm \times 1cm$ per pixel which allows for diverse images from many candidate locations. We check collision for each candidate location using a bounding box with dimensions matching the Fetch robot.

\subsubsection{Autoencoder}
For each environments' autoencoder dataset, we randomly sample 120$K$ collision-free locations from the generated map and capture RGBD images at those locations in the simulator. We use a 0.9\,/\,0.1 train\,/\,test split and the autoencoder model is trained for 200 epochs which on average takes 12 GPU hours. 

\subsubsection{Policy} Using Dijktra's algorithm with costmap optimization, we are able to generate collision-free expert trajectories. In order to map the trajectory into a sequence of \{${forward, right, left}$\} commands, we step through each waypoint position and use a simple discretization strategy. We look ahead 25 steps to determine whether to turn the robot left or right. The robot turns left or right whether the look-ahead position is turned at an offset greater than \ang{20} and outputs turns in increments of \ang{10}. Using only the next position would result in the robot constantly recalculating its direction and the robot would turn at every step to face a new direction. Forward commands are given if the robot is farther from the next position than $0.1m$, and if so moves the robot forward in increments of $0.1m$ until under the threshold. 

Once these commands are generated, we execute the trajectory in the Gibson simulator capturing the RGBD view along every step. This results in a large supervised learning dataset of varying trajectory lengths that we could use for training the agent to learn the policy. We use the trained autoencoder to generate embeddings of RGBD images at each step in the expert trajectories. The average number of steps per trajectory varies from 41 (\texttt{house2}) to 332 (\texttt{area1}). Each training\,/\,testing example is constructed by taking the embeddings from the past 4 steps concatenated with the embedding at the current step. Because larger environments tend to have longer trajectories, we generate 3000 trajectories for \texttt{area1} and \texttt{area2}, 5000 trajectories for \texttt{house17}, and 7000 trajectories for \texttt{house1} and \texttt{house2} to keep the total number of training examples roughly the same. We use $80\%$ of the trajectories for training and the rest for testing. The policy network is trained for 200 epochs which takes around 90 GPU hours. The average accuracy for the policy model is 0.91.

\subsubsection{Goal Checker}
For each environment, we collect positive training examples by randomly sampling 150$K$ positions in the environment for the panoramic goal images and then sample another position within a 0.1$m$ radius for the current image. We collect negative examples by randomly sampling 150$K$ positions for the panoramic goal images and then sample another position at least $1m$ away from the current image. We use a 0.9\,/\,0.1 train\,/\,test split and train the network for 300 epochs. It takes 36 GPU hours and the average accuracy is 0.95 across all environments.


\subsection{Comparison Methods}
We compare our navigation pipeline with an RGBD SLAM~\cite{labbe2019rtab} approach and the target-driven deep RL method from~\cite{THOR}. The methods we examine are described below.

\begin{enumerate}[label=\alph*)]
    \item \textbf{RTABMap} is the Real-Time Appearance-Based Mapping (RTABMap) library provided by~\cite{labbe2019rtab}, which is an RGBD Graph-Based SLAM approach based on an incremental appearance-based loop closure detector. The robot is given the map beforehand.
    \item \textbf{Siamese Actor-Critic (SAC)} is the method proposed by~\cite{THOR}. We only keep one scene specific layer and train the whole network for each environment as we focus on the generalization to different targets under the same environment. We provide a goal-reaching reward of $10.0$ upon task completion and a small penalty of $-0.01$ at each time step. We train it on 100 targets with a maximum step size of 10000 for each episode. We give each environment a budget of 20$M$ frames (steps).
    \item \textbf{Direct Future Prediction (DFP)} algorithm \cite{dosovitskiy2016learning}. This method does not learn to maximize future rewards, but predicts future inputs in the form of RGBD. The actions are learned to maximize the likelihood a future image will align with the predicted. This is the same experimental setup as \cite{minos}. 
    \item \textbf{RL (PPO)} is an agent trained with reinforcement learning, specifically proximal policy optimization [25]. Our agent specifically includes RGBD input. The model consists of a CNN that produces an embedding for visual input, which together with the relative goal vector is used by an actor (GRU) and a critic (linear layer). The experimental setup is the same as specified in ~\cite{habitat}. 
    \item \textbf{Ours (LOC)} is a variant of our proposed pipeline. Instead of using the learned goal checking model, it uses the simulated localized position information and the provided goal coordinate to check whether the agent is at the goal.
    \item \textbf{Ours (no LOC)} is our proposed pipeline with the learned goal checker, without localization, odometry or goal coordinate provided.
\end{enumerate}

a), b), c), d), and e) assume the agent has an idealized localized indoor position and is provided with the \textit{static} goal coordinate as in~\cite{habitat}. The LOC vs. no LOC is meant to be an ablation study comparing what would happen if the goal checker were 100\% accurate. As a result, the agent is able to compute the relative position of the target at each time step and can use this information to check if the goal has been reached.



\subsection{Evaluation Criteria} \label{sec:metrics}
We evaluate the performance of our model using 400 randomly sampled start-goal pairs for which we make sure that a valid path exists. The start and goal locations have never appeared in the training examples. We start the agent at the starting position and provide it with 8 panoramic goal images taken at the goal location. The objective is to navigate to the goal position autonomously with the shortest path possible using only visual input. Our success tolerance is a $0.5m$ radius within the target position. Unlike previous works which do not penalize collision through training and allow collision at runtime~\cite{THOR, habitat}, we simulate real physics and consider the trial a failure when collision occurs.

Similar to many previous works on navigation benchmarks~\cite{THOR, habitat, splmetric, mishkin2019benchmarking}, we focus on three evaluation metrics:
\begin{enumerate}
    \item \textbf{Success Rate} is the number of successful trials over the total number of trials.
    \item \textbf{Success Weighted by Path Length} ($SPL$)~\cite{splmetric} metric is shown in Formula~\ref{spl}, where $l_i$ is the length of the shortest path between start and goal position, $p_i$ is the length of the observed path taken by an agent, and $S_i$ is a binary indicator of success in trial $i$. This metric weighs each success by the quality of path and thus is always $\leq$ \textit{Success Rate}.
    \begin{equation} \label{spl}
        SPL = \frac{1}{N} \sum_{i=1}^{N} S_i \frac{l_i}{\max (p_i, l_i)}
    \end{equation}
    \item \textbf{Observed over Optimal Ratio} ($OOR$) measures the average ratio of observed path length over optimal path length for successful trials.
\end{enumerate}

\subsection{Navigation Results}
Our proposed navigation pipeline significantly outperforms RTABMap and the state-of-the-art deep RL methods in terms of path quality and success rate, as shown in Table~\ref{table:comparison_results}. See Figure~\ref{fig:area1_demo} for several example trajectories generated by our method in the \texttt{area1} environment.

RTABMap~\cite{labbe2019rtab} struggles to localize itself using RGBD alone, due to the high complexity of our testing environments. It succeeds only when the start position is close to the goal position. SAC~\cite{THOR} performs much worse than ours due to the sparse reward and limited number of training frames. In~\cite{THOR}, each environment is a single room and they use synthetic images but our environment can be up to 1031$m^2$ with 40 rooms and we are handling real-world images. Our environment has higher complexity with more obstacles and the entrances to the rooms can be extremely narrow resulting in a difficult solution. SAC needs millions of frames to converge in our environment which is not practical. \cite{THOR} claims they can generalize to new targets by evaluating only on 10 targets that are several steps away from the training targets. In our experimental setup the targets can be anywhere on the map. A majority of the successes for SAC occurred when the target location happens to be in the same room as the start location. Our method also requires much fewer simulation steps\,/\,training frames ($\sim700K$ compared to 20$M$) and less training time (90 GPU hours compared to 300 GPU hours). Additionally DFP~\cite{dosovitskiy2016learning} and RL (PPO)~\cite{habitat} both perform substantially worse than our method. They are purported as working well for examples within similar environments but they fail to generalize well to the unseen targets within the same environment. 

Our method with no LOC achieves similar performance to our variant with LOC. As an ablation study, instead of having a separate goal checking model, a \textit{done} action is generated directly from the policy model, and we find that using a separate goal checking model increases the success rate by $0.2 \sim 0.5$. In the cases where the policy model incorrectly identifies \textit{done}, it either outputs \textit{done} prematurely or passes the goal without terminating. We intentionally keep the amount of training data roughly the same across all environments to evaluate how the performance changes with the complexity of the environment. When the number of rooms is over 30, our method starts to struggle to get to the goal. Despite the reduction in performance due to environmental complexity, our method performs on average $0.556$, $0.585$, $0.148$, and $0.442$ better in success rate than RTABMap, DFP, RL (PPO), and SAC respectively. Given that we achieve high accuracy in smaller environments, we believe the performance in \texttt{area1} and \texttt{area2} will go up if trained on more expert trajectories. A future direction is to analyze the amount of training data needed for a given environmental complexity. For  more information about the performance of our agents in each of the environments see Table \ref{table:our_performance}. 

\begin{table}
    \vspace{2mm}
    \centering {
    \resizebox{\linewidth}{!}{
        \begin{tabular}{cccccccc}
            \toprule
            \textbf{Environment} & \textbf{Model} & \footnotesize{\textit{Success Rate}} & \textit{SPL} & \textit{OOR} \\
            \midrule
            
            \multirow{2}{*}{\makecell{house2 \\ (66$m^2$, 6 rooms)}}
            & Ours (LOC)        & 0.9950 & 0.9810 & 1.066 \\
            & Ours (no LOC)        & 0.9875 & 0.9724 & 1.053 \\
            \midrule
            
            \multirow{2}{*}{\makecell{house1 \\ (89$m^2$, 10 rooms)}} 
            & Ours (LOC)        & 0.9975 & 0.9811 & 1.064 \\
            & Ours (no LOC)        & 0.9225 & 0.8748 & 1.252 \\
            \midrule
            
            \multirow{2}{*}{\makecell{house17 \\ (220$m^2$, 14 rooms)}} 
            & Ours (LOC)        & 0.9800 & 0.7853 & 2.020 \\
            & Ours (no LOC)        & 0.9150 & 0.7179 & 2.389 \\
            \midrule
            
            \multirow{2}{*}{\makecell{area2 \\ (1031$m^2$, 31 rooms)}} 
            & Ours (LOC)        & 0.7250 & 0.5536 & 2.504 \\
            & Ours (no LOC)        & 0.6625 & 0.4714 & 2.948 \\
            \midrule
            
            \multirow{2}{*}{\makecell{area1 \\ (786$m^2$, 40 rooms)}} 
            & Ours (LOC)           & 0.6600 & 0.3954 & 4.504 \\
            & Ours (no LOC)       & 0.5750 & 0.2705 & 5.896 \\
            \midrule
            
            
            \multirow{2}{*}{\makecell{\textit{Average} \\ (483.4$m^2$, 20.2 rooms)}} 
            & Ours (LOC)        & 0.8715 & 0.7393 & 2.232 \\
            & Ours (no LOC)       & 0.8125 & 0.6614 & 2.707 \\
            \bottomrule
        \end{tabular}}}
    \caption{Different method results over 5 environments, with the best values in bold. For SPL higher values are better. For OOR lower values are better. Our method with no LOC achieves similar performance to our variant with LOC. The results with LOC show how the policy performs with a goal checker that has no errors. }
    \label{table:our_performance}
\end{table}

\begin{table}
    \centering {
    \resizebox{\linewidth}{!}{
        \begin{tabular}{cccccc}
            \toprule
            \textbf{Model} & \footnotesize{\textit{Success Rate}} & \textit{SPL} & \textit{OOR}\\
            \midrule

            RTABMap         & 0.2570 & 0.1746 & 25.89 \\
            SAC          & 0.3705 & 0.2363 & 4.076\\
            RL (PPO)     &  0.6650 & 0.5600 & 3.753 \\
            DFP          &  0.2275 & 0.1642 & 8.853 \\
            Ours (no LOC)       & \textbf{0.8125} & \textbf{0.6614} & \textbf{2.707} \\
            \bottomrule
        \end{tabular}}}
    \caption{Comparison of each method's respective performance with the best values in bold. This is the average performance across all environments. Our method outperforms alternative methods in terms of success rate, SPL, and OOR. }
    \label{table:comparison_results}
\end{table}



\section{Conclusions}
We proposed a navigation pipeline which does not rely on odometry, map, compass or indoor position at runtime and is purely based on the visual input and a novel 8-image panoramic goal. Our method learns from expert trajectories generated using RGBD maps of several real environments. Using robotic simulators with real data and photo-realistic rendering, we are able to efficiently collect a large amount of expert trajectories and train in simulation with real-world data, relieving the need of sim-to-real domain adaptation. Our experiments show that the proposed method 1) achieves better performance than cutting edge baselines, especially in complex environments with difficult and long-range path solutions; 2) requires fewer training samples and less training time; and 3) can work across different environments given an RGBD map. Our future work will explore using semantic labels as features for learning navigation, and will test in more complex environments.

\newpage
\bibliographystyle{IEEEtran}
\bibliography{IEEEabrv,mybib}

\end{document}